\documentclass[preprints,article,accept,moreauthors,pdftex]{Definitions/mdpi} 

\firstpage{1} 
\pubvolume{1}
\issuenum{1}
\articlenumber{0}
\pubyear{2021}
\copyrightyear{2020}
\datereceived{} 
\dateaccepted{} 
\datepublished{} 
\hreflink{https://doi.org/} 

\usepackage{amsfonts,amssymb}
\usepackage{wrapfig}
\usepackage{subfig}
\usepackage{multirow}

\Title{The Impact of Global Structural Information in Graph Neural Networks Applications}

\TitleCitation{The Impact of Global Structural Information in Graph Neural Networks Applications}

\Author{Davide Buffelli $^{1*}$\orcidA{}, Fabio Vandin $^{1}$\orcidB{}}

\AuthorNames{Davide Buffelli, Fabio Vandin}

\AuthorCitation{Buffelli, D.; Vandin, F.}

\address[1]{$^{1}$ \quad Department of Information Engineering, University of Padova; Padova, Italy}

\corres{Correspondence: davide.buffelli@phd.unipd.it}

\abstract{
Graph Neural Networks (GNNs) rely on the graph structure to define an aggregation strategy where each node updates its representation by combining information from its neighbours. A known limitation of GNNs is that, as the number of layers increases, information gets \textit{smoothed} and \textit{squashed} and node embeddings become indistinguishable, negatively affecting performance. 
Therefore, practical GNN models employ few layers and only leverage the graph structure in terms of limited, small neighbourhoods around each node. Inevitably, practical GNNs do not capture information depending on the global structure of the graph.
While there have been several works studying the limitations and expressivity of GNNs, the question of whether practical applications on graph structured data require global structural knowledge or not, remains unanswered. 
In this work, we empirically address this question by giving access to \textit{global} information to several GNN models, and observing the impact it has on downstream performance.
Our results show that global information can in fact provide significant benefits for common graph-related tasks. We further identify a novel regularization strategy that leads to an average accuracy improvement of more than $5\%$ on all considered tasks.}


\begin{document}
\section{Introduction}
Graph Neural Networks (GNNs) \cite{Wu2019ACS} are deep learning models for graph structured data, which achieve state-of-the-art results for many graph-related tasks.
Most popular GNNs fall into the message-passing framework \citep{Gilmer2017}, and are denoted as Message Passing Neural Networks (MPNNs)\footnote{In this paper we use the terms GNN and MPNN interchangeably.}.
MPNNs have become increasingly popular thanks to their simplicity, extensibility, and empirical effectiveness. MPNNs adopt a message passing mechanism where, at each layer, every node receives a message from its 1-hop neighbours. The incoming messages for each node are aggregated in a permutation-invariant fashion and used to update the node's representation by the means of a learnable function (usually implemented with a neural network). The final node representations (also referred to as \emph{node embeddings}) are then used to perform some graph-related downstream task, for example graph classification or node classification. 
Empirically, the best results are obtained when the message passing procedure is repeated a relatively small number of times (typical numbers are 2 to 5), as a higher number of layers leads to over-smoothing \cite{Li2018DeeperII} and over-squashing \cite{alon2021on}. Thus, \textit{practical} GNNs are only leveraging the graph structure in the form of small neighbourhoods around each node. A direct consequence of this limitation is that GNNs are not capable of accessing, or extracting, information that depends on the whole structure of the graph (e.g. random walk probabilities \cite{MASUDA20171}). 

In this work we are interested in studying the consequences of the over-smoothing and over-squashing issues.
In more detail, we are interested in understanding whether global information (i.e., information that depends on the \textit{whole} structure of the graph, and that cannot be recovered by just focusing on local neighbourhoods) is important for GNNs and their practical applications. 

In fact, there is an ongoing debate in the GNN research community on whether it is needed to have ``deep'' GNNs \cite{bronstein_tds}, or, if most tasks of interest only require access to local neighbourhoods. We tackle this question directly at its root, and address the overlooked aspect of whether \textit{global} structural information is useful for GNN models, by studying if \textit{global} structural information is important in practical scenarios.
In more detail, we introduce three different ways to provide GNN models with \textit{global} structural information, and study how they affect the performance of state-of-the-art MPNNs on common graph related tasks. The three strategies to include \textit{global} structural information we consider are: (i) providing the model direct access to the adjacency matrix, (ii) providing the model direct access to random walks with restart coefficients, and (iii) combining (ii) with a regularization term which enforces the role of the information extracted by random walks with restart. These methods are introduced to study the impact of global information, and are not meant to be used as practical strategies to improve the performance of GNNs. On the latter aspect, we show that the sole use of our regularization term provides significant gains in performance while being easily and efficiently applicable to any GNN model. The use of random walks with restart is also supported by a theoretical contribution which proves they can increase the ability of GNNs in distinguishing non-isomorphic graphs.





\paragraph{\textbf{Our Contribution.}}
Previous studies on the capabilities and limitations of GNNs have focused on the relation between GNNs and the Weisfeiler-Leman (WL) algorithm \cite{wl1968} to study the \textit{theoretical} expressiveness of these models (e.g. \cite{Morris2019}), or on how to alleviate the over-smoothing and over-squashing issues (e.g. \cite{Li2018DeeperII,li2021gnn1000,alon2021on}). There are however no empirical studies on the practical impact of \textit{global} information (i.e., information that depends on the \textit{whole} structure of the graph) in MPNNs.

We assess whether providing \textit{global} information regarding the whole graph structure has a significant impact on the performance of state-of-the-art MPNNs. 
In this regard, our contributions are threefold. 
\begin{itemize}
\item We propose and formalize three different types of \textit{global} structural information ``injection''. We test how the injection of \textit{global} structural information impacts the performance of 6 GNN models on both transductive and inductive tasks. Results show that the injection of \textit{global} structural information significantly impacts current state-of-the-art models on common graph-related tasks.
\item As we discuss later in the paper, injecting \textit{global} structural information can be impractical. We then identify a novel and practical regularization strategy, called RWRReg, based on random walks with restart \cite{Pageetal98}. RWRReg maintains the permutation-invariance of GNN models, and leads to an average $5\%$ increase in accuracy on both node classification, and graph classification. 
\item We introduce a theoretical result proving that the information extracted by random walks with restart can ``speed up'' the 1-Weisfeiler-Leman (1-WL) algorithm \cite{wl1968}. In more detail we show that by constructing an initial coloring based on random walks with restart probabilities, the 1-WL algorithm always terminates in one iteration. Given the known relationship between GNNs and the 1-WL algorithm, this result shows that providing information obtained from random walks with restart to GNN models can improve their \textit{practical} ability of distinguishing non-isomorphic graphs.
\end{itemize}

\section{Preliminaries}
In this section we introduce the notation we use throughout the paper, and provide a brief introduction to GNNs and random walks with restart (RWR; also known as Personalized PageRank \cite{Pageetal98}).

\subsection{Notation}
We use uppercase bold letters for matrices ($\boldsymbol{M}$), and lowercase bold letters for vectors ($\boldsymbol{v}$). We use plain letters with subscript indices to refer to a specific element of a matrix ($M_{i,j}$), or of a vector ($v_{i}$). 
We refer to the vector containing the $i$-th row of a matrix with the subscript ``$i, :$'' ($\boldsymbol{M}_{i, :}$), while we refer to the $i$-th column with the subscript ``$:, i$'' ($\boldsymbol{M}_{:, i}$). 

A graph $\mathcal{G} = (\mathcal{V}, E)$, where $\mathcal{V} = \{ 1, .., n \}$ is the set of nodes and $E \subseteq \mathcal{V} \times \mathcal{V}$ is the set of edges, is represented by a tuple $(\boldsymbol{X}, \boldsymbol{A})$. $\boldsymbol{X}$ is an $n \times d$ matrix where the $i$-th row contains the $d$-dimensional feature vector of the $i$-th node, and $\boldsymbol{A}$ is the $n \times n$ adjacency matrix. For the sake of clarity we restrict our presentation to undirected graphs, but similar concepts can be applied to directed graphs.

\subsection{Graph Neural Networks}
In graph representation learning, the goal is to learn a vector representation (also referred to as \emph{node embedding}) for each node that can then be used to effectively perform downstream tasks.
The message-passing framework \citep{Gilmer2017}, to which most GNNs belong, is based on the following procedure: each node receives messages from its neighbours,  \textit{aggregates} them, and \textit{updates} its representation based on the aggregated messages and its previous representation.
For a node $v$, with neighbours $\mathcal{N}_v$, we can represent the operations at the $\ell$-th layer of message-passing as follows:
\begin{align*}
\mathbf{m}^{(v, \ell)}  &= \text{AGGREGATE}(\{ \mathbf{H}_{u}^{(\ell)} \text{ } \forall u \in \mathcal{N}_v \}) \\
\mathbf{H}_{v}^{(\ell+1)} &= \text{UPDATE}(\mathbf{H}_{v}^{(\ell)}, \mathbf{m}^{(v, \ell)})   
\end{align*}
where $\mathbf{H}^{(\ell)}$ is a matrix where the $i$-th row contains the representation of node $i$ at layer $\ell$, $\text{AGGREGATE}$ is a permutation invariant function (e.g., average or sum) that takes as input the set of representations of the neighbours and aggregates them into a message $\mathbf{m}$, and $\text{UPDATE}$ is usually a learnable function implemented with a neural network. The initial representation (at layer 0) is defined as $\mathbf{H}^{(0)}=\mathbf{X}$. As such, after $k$ message-passing iterations, the representation of a node $v$ depends on its $k$-hop neighbourhood (i.e., all the nodes at distance at most $k$ from $v$). 
The GNNs proposed in literature differ on how they implement the AGGREGATE and UPDATE functions \citep{Wu2019ACS,kipf2017semi,hamilton2017inductive,velickovic2018graph}.

\subsection{Random Walk's with Restart}
\label{S:theory}
A RWR \cite{Pageetal98} for node $i$ returns a vector $\boldsymbol{r}^{(i)}$ of size $n$ which satisfies the following equation:
\[
\boldsymbol{r}^{(i)} = (1-c)\boldsymbol{W}\boldsymbol{r}^{(i)} + c\boldsymbol{e}^{(i)}
\]
where $\boldsymbol{e}^{(i)}$ is a vector where the $i$-th element is 1 and all the others are 0, $c$ is the restart probability, and $\boldsymbol{W}$ is the transition matrix of the random walk. The restart probability $c$ defines the probability that the walk ``jumps'' back to the starting node (a common value for $c$, used in many libraries, is $0.15$). The RWR vector can be computed using the power iteration method, and over the year a large number of methods have been developed for its efficient and practical computation, or approximation, even for large scale graphs (e.g., \citep{Lofgren2015EfficientAF,Tong2006FastRW}). Elements of $\boldsymbol{r}^{(i)}$ capture the relative relationships between nodes \cite{Tong2006FastRW}, and the 
RWR vectors capture the global structure of the graph \cite{Jin2019SupervisedAE,10.1145/1027527.1027531}.

\section{Random Walks with Restart and the Weisfeiler-Leman Algorithm}
We provide analytical evidence that RWR can significantly empower MPNNs by proving a connection with the 1-Weisfeiler-Leman (1-WL) algorithm \cite{wl1968}.

The 1-WL algorithm is a well known method for testing the isomorphism of two graphs. The 1-WL algorithm uses an iterative coloring, or relabeling, scheme, in which all nodes are initially assigned the same label (e.g., the value $1$). It then iteratively refines the color of each node by aggregating the multiset of colors in its neighborhood with the use of a hash function. At every iteration, the feature representation of a graph is the histogram of resulting node colors. If, at a certain iteration of this process, two graphs have a different feature representation, then the two graphs are not isomorphic. (For a more detailed description of the 1-WL algorithm we refer the reader to~\citep{wl1968,shervashidze2011weisfeiler}.)

It is known that not all non-isomorphic graphs are distinguishable by the 1-WL algorithm, and that $n$ iterations are enough to distinguish two graphs of $n$ vertices which are distinguishable by the 1-WL algorithm. There is a tight connection between 1-WL and MPNNs \cite{kipf2017semi, Xu2018HowPA}. In particular, graphs that can be distinguished in $k$ iterations by the 1-WL algorithm, can be distinguished by \textit{certain} GNNs in $k$ message passing iterations \cite{Morris2019}. This implies that  when using a GNN that can theoretically achieve the distinguishing power of the 1-WL algorithm, if such GNN is deployed with $k^\prime$ layers, it will not be able to distinguish graphs that are distinguishable by the 1-WL algorithm with $k^{\prime\prime} > k^\prime$ iterations.

Here, we prove that graphs that are distinguishable by 1-WL in $k$ iterations have different feature representations extracted by RWR of length $k$, and hence if we use the RWR feature representations as initial coloring for the 1-WL algorithm, then the algorithm will always finish in one iteration. 
Given a graph $G=(V,E)$, we define its \emph{$k$-step RWR representation} as the set of vectors $\mathbf{r}_v  = [r_{v,u_1}, \dots, r_{v,u_n}]$, $v\in V$, where each entry $r_{v,u}$ is the probability that a RWR of length $k$ starting in $v$ ends in $u \in V$.

\begin{Proposition}
Let $G_1=(V_1,E_1)$ and $G_2=(V_2,E_2)$ be two non-isomorphic graphs for which the 1-WL algorithm terminates with the correct answer after $k$ iterations and starting from the labelling of all $1$'s. Then the $k$-step RWR representations of $G_1$ and $G_2$ are different.
\end{Proposition}

The proof can be found in Appendix \ref{appendix_proof}.
Since $k$ iterations of the 1-WL algorithm are performed by MPNNs of depth $k$, but in practice MPNNs are limited to small depths, this result shows that RWR can empower MPNNs with relevant information that is discarded in practice.

We further provide an empirical analysis of RWR and their capability of encapsulating global information in Appendix \ref{S:rw_analysis}.

\section{Injecting Global Information in MPNNs}\label{S:injections}
To test if MPNNs are missing on important information that is encoded in the structure of a graph, we inject global structural information into existing MPNN models, and test how the performance of these models changes in several graph-related tasks. Intuitively, based on a model's performance when injected with different types of global structural information, we can understand  
if this additional knowledge can improve performance on the considered tasks. 
In the rest of this section we present the types of global structural information injection that we consider, and the models chosen for our experimental evaluation.

\subsection{Types of Global Structural Information Injection}
We consider three different types of global structural information injection, described below. The injection strategies presented in this section are not designed for \textit{practical} use, as the scope of these strategies is to help us understand the importance of global structural information. At this point, our objective is to study the impact of global structural information that is not accessible to GNN models. We discuss scalability and practical aspects in Section \ref{S:practical}. 

\begin{description}
\item \textbf{Adjacency Matrix.} We provide GNNs with direct access to the adjacency matrix by concatenating each node's adjacency matrix row to its feature vector. This explicitly empowers the GNN model with the connectivity of each node, and allows for higher level structural reasoning when considering a neighbourhood (the model will have access to the connectivity of the whole neighbourhood when aggregating messages from neighbouring nodes).

\item \textbf{Random Walk with Restart (RWR) Matrix.} We perform RWR \citep{Pageetal98} from each node  $v$, thus obtaining a $n$-dimensional vector that gives a score of how much $v$ is ``related" to every other node in the graph. For every node, we concatenate its vector of RWR coefficients to its feature vector. The choice of RWR is motivated by their capability to capture the relevance between two nodes \citep{Tong2006FastRW} and the global structure of a graph \cite{Jin2019SupervisedAE,10.1145/1027527.1027531}, and by the possibility to modulate the exploration of long-range dependencies by changing the restart probability. Intuitively, if a RWR starting at node $v$ is very likely to visit a node $u$ (e.g., there are multiple paths that connect the two), then there will be a high score in the RWR vector for $v$ at position $u$. This gives the GNN model higher level information about the global structure of the graph, and, again, it allows for high level reasoning on neighbourhood connectivity.

\item \textbf{RWR Matrix + RWR Regularization.} Together with the addition of the RWR score vector to the feature vector of each node, we also introduce a regularization term based on RWR that pushes nodes with mutually high RWR scores to have embeddings that are close to each other (independently of how far they are in the graph).
Let $\boldsymbol{S}$ be the $n \times n$ matrix with the RWR scores. We define the RWRReg (\underline{R}andom \underline{W}alk  with \underline{R}estart \underline{Reg}ularization) loss as follows:
\[ \mathcal{L}_{\text{\textit{RWRReg}}} =  \sum_{i,j\in V} S_{i,j} \lvert \lvert \boldsymbol{H}_{i, :} - \boldsymbol{H}_{j, :} \rvert \rvert^{2} \]
where $\boldsymbol{H}$ is a matrix of size $n \times d$ containing $d$-dimensional node embeddings that are in between message-passing layers (see Appendix \ref{Appendix_model_details} for the exact point in which $\boldsymbol{H}$ is considered for each model). With this approach, the loss function used to train the model becomes: $  \mathcal{L} =  \mathcal{L}_{\text{\textit{original}}} + \lambda \mathcal{L}_{\text{\textit{RWRReg}}}$, where $\mathcal{L}_{\text{\textit{original}}}$ is the original loss function for each model, and $\lambda$ is a balancing term. 
In Appendix \ref{appendix:compute_rwr} we show how to compute the RWRReg term efficiently using GPUs.
We expect this type of information injection to have the highest impact on performance of the models on downstream tasks.
\end{description}

\subsection{Choice of Models}\label{S:models}
In order to test the effect of the different types of global structural information injection and to obtain results that are indicative of the whole class of MPNNs models, we conceptually identify four different categories of MPNNs from which we select representative models.
\paragraph{Simple Aggregation Models.} Such models utilize a ``simple'' aggregation strategy, where each node receives messages (e.g., feature vectors) from its neighbours, aggregates them by assigning the same ``importance'' to each neighbour (e.g., by averaging their messages), and uses the aggregated messages to update its embedding vector. As a representative we choose GCN~\citep{kipf2017semi}, one of the fundamental and widely used GNNs models. 
We also consider GraphSage~\citep{hamilton2017inductive}, as it represents a different computation strategy where a set of neighborhood aggregation functions are learned, and a sampling approach is used for defining fixed size neighbourhoods.
\paragraph{Attention Models.} Several models have used an attention mechanism in a GNN scenario \citep{Lee2018AttentionMI,Lee2018GraphCU,velickovic2018graph,Zhang2018GaANGA}. These methods differ from the previous category as they use an attention mechanism to assign a different ``weight'', or ``importance'', to each neighbour. As a representative we focus on GAT~\citep{velickovic2018graph}, the first to present an attention mechanism over nodes for the aggregation phase, and one of the best performing models on several datasets. Furthermore, it can be used in an inductive scenario.
\paragraph{Pooling Techniques.} Pooling on graphs is a very challenging task, since it has to take into account the underlying graph structure. At a high level, pooling methods provide a coarsened version of the input graph by combining groups of nodes into clusters. Among the methods that have been proposed for differentiable pooling on graphs~\citep{cangea2018,Ying2018HierarchicalGR,diehl2019,gao2019graph,Lee2019SelfAttentionGP}, we choose DiffPool~\citep{Ying2018HierarchicalGR} for its strong empirical results. Furthermore, it can learn to dynamically adjust the number of clusters (the number is a hyperparameter, but the network can learn to use fewer clusters if necessary). 
\paragraph{Beyond WL.} \citet{Morris2019} prove that message-passing GNNs cannot be more powerful than the 1-WL algorithm, and propose $k$-GNNs, which rely on a \textit{subgraph message-passing} mechanism and are proven to be as powerful as the $k$-WL algorithm. Another approach that goes beyond the WL algorithm was proposed by \citet{Murphy2019RelationalPF}. Both models are computationally intractable in their initial theoretical formulation, so approximations are needed. As representative we choose $k$-GNNs, to test if subgraph message-passing is affected by additional global structural information.

\section{Evaluation of the Injection of Global Structural Information\label{S:eval}}
We now present our framework for evaluating the effects of the injection of global structural information into GNNs, and the results of our experiments \footnote{The source code used for our experiments is provided as Supplementary Material.}. 
We consider one \textit{transductive} task (node classification) and two \textit{inductive} tasks (graph classification, and triangle counting).
We use each architecture for the task that better suits its design: GCN, GraphSage, and GAT for node classification, and DiffPool and $k$-GNN for graph classification. We add an adapted version of GCN for graph classification, as a common strategy for this task is to deploy a node-level GNN, and then apply a \textit{readout} function to combine node embeddings into a global graph embedding vector.

With regards to datasets, for node classification we considered the three most used benchmarking datasets in literature: Cora, Citeseer, and Pubmed \citep{Sen_2008}. Analogously, for graph classification we chose three frequently used datasets: ENZYMES, PROTEINS, and D\&D \citep{KKMMN2016}. Dataset statistics can be found in Appendix \ref{appendix_Dataset_statistics}.

For all the considered models we take the hyperparameters from the implementations released by the authors. The only parameter tuned using the validation set is the balancing term $\lambda$ when RWRReg is applied. We found that the RWRReg loss tends to be larger than the Cross Entropy loss for prediction, and the best values for $\lambda$ lie in the range $[10^{-9}, 10^{-6}]$. For all the RWR-based techniques we used a restart probability of $0.15$\footnote{We use 0.15 as it is a common default value used in many papers and software libraries.}. (The effects of different restart probabilities are explored in Section \ref{S:practical}.)
Detailed information on our implementations can be found in Appendix \ref{Appendix_model_details}.

\paragraph{Node Classification.}
\begin{table}
\caption{\label{tab:node_classification}Node classification accuracy results of different models with added Adjacency matrix features (AD), RWR features (RWR), and RWR features + RWR Regularization (RWR+RWRReg).}
\begin{center}
\begin{tabular}{l  l c c c c c}
\toprule
 \textbf{Model} & \textbf{Structural} & \multicolumn{3}{c}{\textbf{Dataset}} \\
 &\textbf{Information}  & Cora & Pubmed & Citeseer\\
 \midrule
         &  none   & $0.799 \pm 0.029$ & $0.776 \pm 0.022$ & $0.663 \pm 0.095$\\ 
         & AD       & $0.806 \pm 0.035$ & $0.779 \pm 0.070$ & $0.653 \pm 0.104$\\  
 GCN& RWR      & $0.817 \pm 0.025$ & $0.782 \pm 0.042$ & $0.665 \pm 0.098$\\
         & RWR+RWRReg & $\boldsymbol{0.842 \pm 0.026}$ & $\boldsymbol{0.811 \pm 0.037}$ & $\boldsymbol{0.690 \pm 0.102}$\\	
\midrule
                    &  none   & $0.806 \pm 0.017$ & $0.807 \pm 0.016$ & $0.681 \pm 0.021$\\ 
                    & AD       & $0.803 \pm 0.014$ & $0.803 \pm 0.013$ & $0.688 \pm 0.020$\\  
 GraphSage& RWR      & $0.816 \pm 0.014$ & $0.807 \pm 0.015$ & $0.693 \pm 0.019$\\
                   & RWR+RWRReg & $\boldsymbol{0.837 \pm 0.015}$  & $\boldsymbol{0.820 \pm 0.010}$  & $\boldsymbol{0.728 \pm 0.020}$\\	
\midrule
        &  none   & $0.815 \pm 0.021$ & $0.804 \pm 0.011$ & $0.664 \pm 0.008$\\ 
        & AD       & $0.823 \pm 0.019$ & $0.796 \pm 0.014$ & $0.672 \pm 0.017$\\  
 GAT& RWR      & $0.833 \pm 0.020$ & $0.811 \pm 0.009$ & $0.686 \pm 0.009$\\
        & RWR+RWRReg   & $\boldsymbol{0.848 \pm 0.019}$  & $\boldsymbol{0.828 \pm 0.010}$  & $\boldsymbol{0.701 \pm 0.011}$\\	          
 \bottomrule 
\end{tabular}
\end{center}
\end{table}
For each dataset we follow the approach that has been widely adopted in literature: we take 20 labeled nodes per class as training set, 500 nodes as validation set, and 1000 nodes for testing. Most authors have used the train/validation/test split defined by \citet{Yang2016RevisitingSL}. Since we want to test the general effect of the injection of global structural information, we differ from this approach and we do not rely on a single split. We perform 100 runs, where at each run we randomly sample 20 nodes per class for training, 500 random nodes for validation, and 1000 random nodes for testing. We then report mean and standard deviation for the accuracy on the test set over these 100 runs.

Results are summarized in Table \ref{tab:node_classification}, where we observe that the simple addition of RWR features to the feature vector of each node is sufficient to give a performance gain (up to 2\%). The RWRReg term then significantly increments the gain (up to \textbf{7.5\%}). These results show that, perhaps surprisingly, even for the task of node classification global structural information is important. 

\paragraph{Graph Classification.}
\begin{table}
\caption{\label{tab:graph_classification}Graph classification accuracy results of different models with added Adjacency matrix features (AD), RWR features (RWR), and RWR features + RWR Regularization (RWR+RWRReg).}
\begin{center}
\begin{tabular}{l  l c c c c}
\toprule
 \textbf{Model} & \textbf{Structural} & \multicolumn{3}{c}{\textbf{Dataset}} \\
 &\textbf{Information} & ENZYMES & D\&D & PROTEINS\\
  \midrule 
              &  none   & $0.570 \pm 0.052$ & $0.755 \pm 0.028$ & $0.740 \pm 0.035$\\ 
              & AD       & $0.591 \pm 0.076$ & $0.779 \pm 0.022$ & $0.775 \pm 0.042$\\  
 GCN     & RWR      & $0.584 \pm 0.055$ & $0.775 \pm 0.023$ & $0.784 \pm 0.034$\\
              & RWR+RWRReg   & $\boldsymbol{0.616 \pm 0.065}$ & $\boldsymbol{0.790 \pm 0.023}$ & $\boldsymbol{0.795 \pm 0.032}$\\	
  \midrule  
              &  none    & $0.661 \pm 0.031$ & $0.793 \pm 0.022$ & $0.813 \pm 0.017$\\ 
              & AD       & $0.711 \pm 0.027$ & $0.837 \pm 0.020$ & $0.821 \pm 0.039$\\  
 DiffPool & RWR      & $0.687 \pm 0.025$ & $0.824 \pm 0.028$ & $0.783 \pm 0.043$\\
              & RWR+RWRReg   & $\boldsymbol{0.721 \pm 0.039}$ & $\boldsymbol{0.840 \pm 0.024}$ & $\boldsymbol{0.834 \pm 0.038}$\\	
 \midrule 
                 & none     & $0.515 \pm 0.111$ & $0.756 \pm 0.021$ & $0.763 \pm 0.043$\\ 
                 & AD       & $0.572 \pm 0.063$ & $0.778 \pm 0.020$ & $0.751 \pm 0.034$\\  
 $k$-GNN & RWR      & $\boldsymbol{0.573 \pm 0.077}$ & $\boldsymbol{0.794 \pm 0.022}$ & $0.781 \pm 0.028$\\
                 & RWR+RWRReg  & $0.571 \pm 0.080$ & $0.786 \pm 0.021$ & $\boldsymbol{0.785 \pm 0.026}$\\	         
 \bottomrule 
\end{tabular}
\end{center}
\end{table}
Following the approach from \citet{Ying2018HierarchicalGR} and \citet{Morris2019} we use 10-fold cross validation, and report mean and standard deviation of the accuracy on graph classification. Results are summarized in Table \ref{tab:graph_classification}. 
The performance gains given by the injection of global structural information are even more apparent than for the node classification task. Intuitively, this is explained by the fact that the global structure of the nodes in a graph is important for distinguishing different graphs. Most notably, the addition of the adjacency features is sufficient to give a large performance boost (up to \textbf{11\%}). 

Surprisingly, models like DiffPool and $k$-GNN show an important difference in accuracy (up to \textbf{10\%}) when there is injection of structural information, meaning that even the most advanced methods suffer from the inability to exploit global structural information. 

\paragraph{Counting Triangles.}  
The TRIANGLES dataset \cite{knyazev2019understanding} is composed of randomly generated graphs, where the task is to count the number of triangles contained in each graph. This is a hard task for GNNs 
and, as in \citet{knyazev2019understanding}, we use node degrees as node features to impose some structural information in the network.
The TRIANGLES dataset has a test set with 10'000 graphs, of which half are similar in size to the ones in the training and validation sets (4-25 nodes), and half are bigger (up to 100 nodes). This permits an evaluation of a model's capabilities generalization to graphs of unseen sizes. 

\begin{wraptable}{r}{6.7cm}
\caption{\label{tab:counting_triangle_mse}Mean Squared Error of GCN with different types of global structural information injection on the TRIANGLES dataset.}
\begin{center}
\resizebox{6.5cm}{!}{%
\begin{tabular}{ l c c c}
 \toprule
 \textbf{Model}  & \multicolumn{3}{c}{\textbf{TRIANGLES Test Set}} \\
& Global & Small & Large \\
  \midrule
 GCN                   & $2.290$ & $1.311$ & $3.608$\\ 
 GCN-AD             & $4.746$ & $1.162$ & $5.971$ \\  
 GCN-RWR            & $2.044$ & $\boldsymbol{1.101}$ & $2.988$ \\
 GCN-RWR+RWRReg       & $\boldsymbol{2.029}$ & $1.166$ & $\boldsymbol{2.893}$ \\	
 \bottomrule
\end{tabular}
}
\end{center}
\end{wraptable}
For this regression task we use a three layer GCN, and we minimize the Mean Squared Error (MSE) loss (more details can be found in Appendix \ref{Appendix_model_details}).  
Table \ref{tab:counting_triangle_mse} presents MSE results on the test dataset as a whole and on the two splits separately. We see that the addition of RWR features and of RWRReg provides significant benefits (up to \textbf{19\%} improvements), specially when the model has to generalize to graphs of unseen sizes, while the addition of adjacency features leads to overfitting.

\section{Practical Aspects}\label{S:practical}

From the results shown in Section \ref{S:eval}, it would be tempting to propose the addition of adjacency matrix information or RWR information into node feature vectors as a strategy to improve the performance of GNN models. However, the benefits introduced by such a strategy come at a high cost: adding $n$ features increases the input size of $n \times n$ elements (which is prohibitive for large graphs). Furthermore, all the considered models have a weight matrix at each layer that depends on the feature dimension, which means we are also increasing the number of parameters at the first layer by $n \times d^{(1)}$ (where $d^{(1)}$ is the dimension of the feature vector for each node after the first GNN layer). In this section we propose a practical way to take advantage of the injection of global structural information without increasing the number of parameters, and controlling the memory consumption during training.

\begin{table}
\caption{\label{tab:rwrreg} Results for the addition of \textit{only} the RWRReg term to existing models on node classification (accuracy), graph classification (accuracy), and triangle counting (MSE-- lower is better).}
\begin{center}
\begin{tabular}{l  l c c c c c}
\toprule
 \textbf{Model} & \textbf{Regularization} & \multicolumn{3}{c}{\textbf{Dataset}} \\
\midrule
\midrule
 & & \multicolumn{3}{c}{\textbf{Node Classification}}\\
 &  & Cora & Pubmed & Citeseer\\
 \midrule
 \multirow{2}{*}{GCN}&  none   & $0.799 \pm 0.029$ & $0.776 \pm 0.022$ & $0.663 \pm 0.095$\\ 
 & RWRReg      & $\boldsymbol{0.861 \pm 0.025}$ & $\boldsymbol{0.799 \pm 0.034}$ & $\boldsymbol{0.686 \pm 0.096}$\\
\midrule
 \multirow{2}{*}{GraphSage}&  none   & $0.806 \pm 0.017$ & $0.807 \pm 0.016$ & $0.681 \pm 0.021$\\ 
 & RWRReg      & $\boldsymbol{0.841 \pm 0.016}$ & $\boldsymbol{0.818 \pm 0.017}$ & $\boldsymbol{0.721 \pm 0.021}$\\
\midrule
 \multirow{2}{*}{GAT}&  none   & $0.815 \pm 0.021$ & $0.804 \pm 0.011$ & $0.664 \pm 0.008$\\ 
 & RWRReg      & $\boldsymbol{0.824\pm 0.022}$ & $\boldsymbol{0.811 \pm 0.013}$ & $\boldsymbol{0.702 \pm 0.013}$\\         
 \midrule
 \midrule 
 & & \multicolumn{3}{c}{\textbf{Graph Classification}}\\
 & & ENZYMES & D\&D & PROTEINS\\
 \midrule
 \multirow{2}{*}{GCN}     &  none   & $0.570 \pm 0.052$ & $0.755 \pm 0.028$ & $0.740 \pm 0.035$\\
      & RWRReg      & $\boldsymbol{0.621 \pm 0.041}$ & $\boldsymbol{0.786 \pm 0.024}$ & $\boldsymbol{0.785 \pm 0.036}$\\
  \midrule  
 \multirow{2}{*}{DiffPool} &  none    & $0.661 \pm 0.031$ & $0.793 \pm 0.022$ & $0.813 \pm 0.017$\\
  & RWRReg      & $\boldsymbol{0.733 \pm 0.032}$ & $\boldsymbol{0.822 \pm 0.025}$ & $\boldsymbol{0.820 \pm 0.038}$\\	
 \midrule 
 \multirow{2}{*}{$k$-GNN} & none     & $0.515 \pm 0.111$ & $0.756 \pm 0.021$ & $0.763 \pm 0.043$\\
  & RWRReg      & $\boldsymbol{0.582 \pm 0.075}$ & $\boldsymbol{0.787 \pm 0.022}$ & $\boldsymbol{0.780 \pm 0.028}$\\ 
 \midrule
 \midrule
 & & \multicolumn{3}{c}{\textbf{Triangles Test Set}}\\
 & & Global & Small & Large \\
 \midrule 
 \multirow{2}{*}{GCN}  & none & $2.290$ & $1.311$ & $3.608$\\ 
   & RWRReg & $\boldsymbol{2.187}$ & $\boldsymbol{1.282}$ & $\boldsymbol{3.014}$ \\
 \bottomrule 
\end{tabular}
\end{center}
\end{table}

\subsection{RWRReg}
From Section \ref{S:eval}, the use of RWR coefficients as additional features coupled with the additional RWRReg term is the strategy that provides the highest performance improvement on all tasks.
As discussed at the beginning of this section, the addition of RWR coefficients can be problematic, and hence we study the impact of using \textbf{only} the RWRReg term. We consider the same settings and tasks presented in Section \ref{S:eval}, and results are shown in Table \ref{tab:rwrreg}. The results show that the sole addition of the RWRReg term increases the performance of the considered models by more than \textbf{5\%}.
At the same time, RWRReg \textbf{(i)} does not increase the input size or the number of parameters, \textbf{(ii)} does not require additional operations at inference time, \textbf{(iii)} does not require additional supervision (it is in fact a \textit{self-supervised} objective), \textbf{(iv)} maintains the permutation invariance of MPNN models, and \textbf{(v)} there is a vast literature on efficient methods for computing RWR, even for web-scale graphs (e.g., \cite{Lofgren2015EfficientAF,Wei_2018,Wang_2019}). Hence, the only downside of RWRReg is the storage of the RWR matrix during training on very large graphs.

\begin{figure}[h!]
\begin{center}
  \subfloat[]{\includegraphics[height=1.4in]{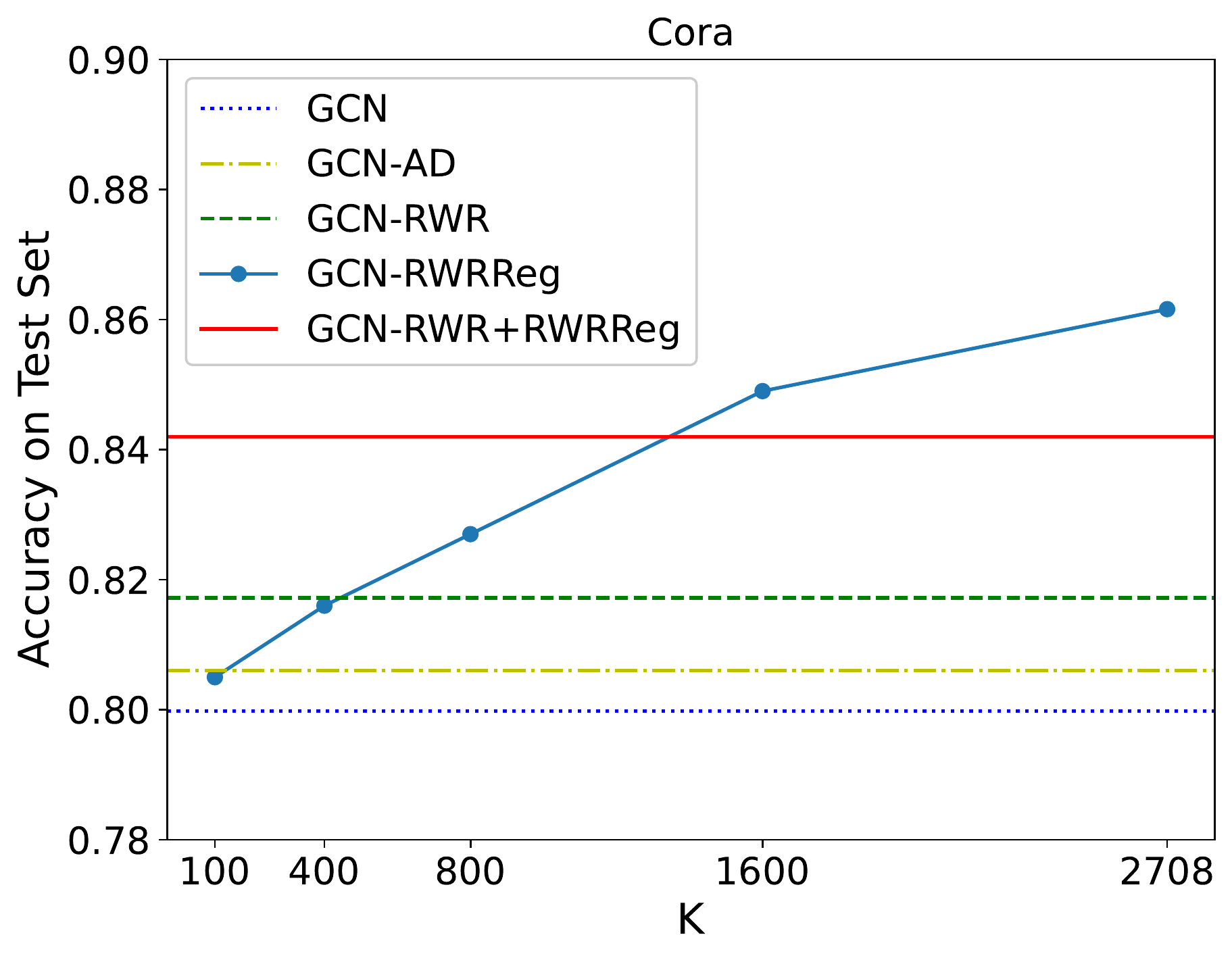}}
  \subfloat[]{\includegraphics[height=1.4in]{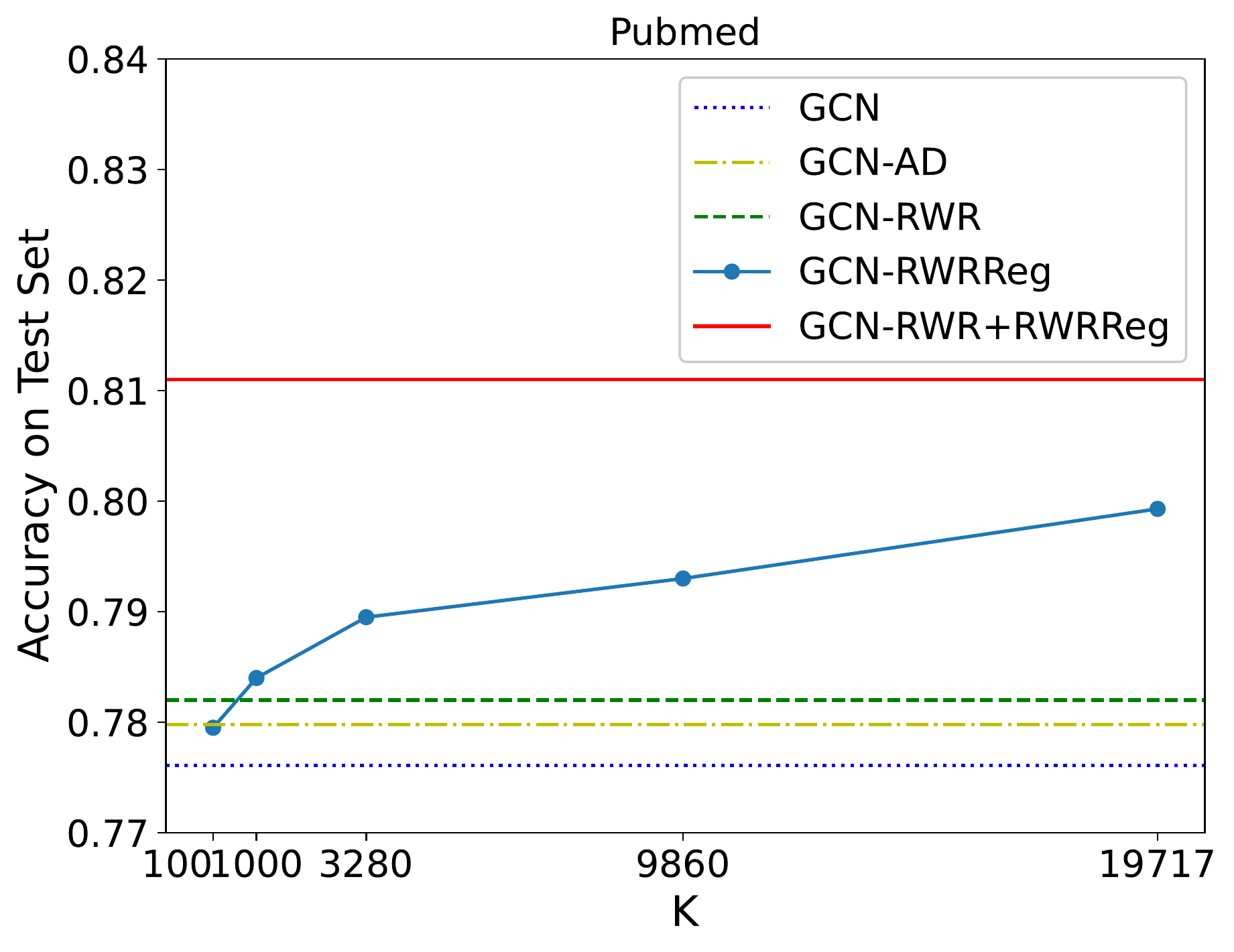}}
  \subfloat[]{\includegraphics[height=1.4in]{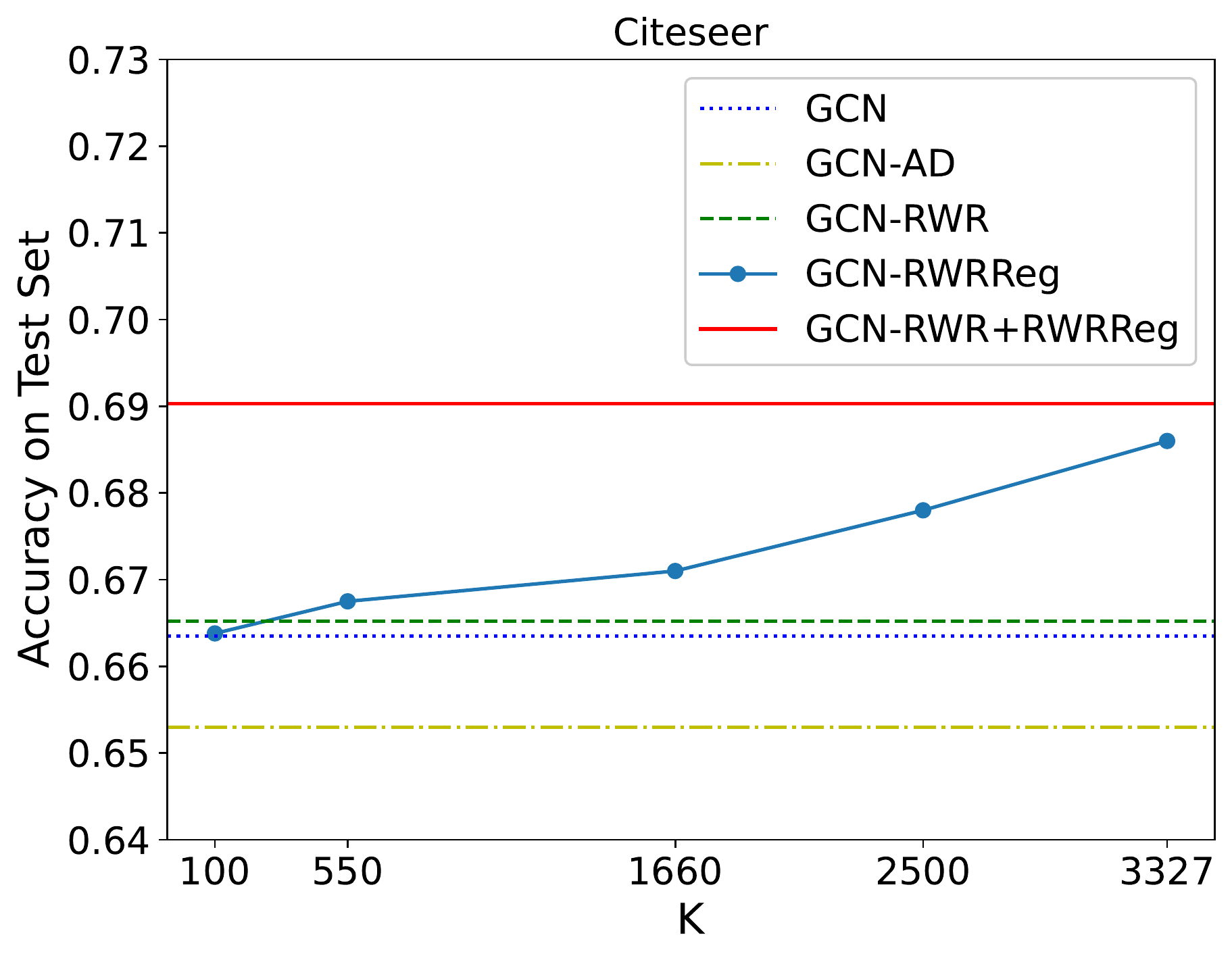}}
\end{center}
  \caption{Performance of GCN on node classification for different values of $K$ when trained with RWRReg with \textit{Top-$K$} sparsification of the RWR matrix.}
  \label{fig:gcn_top_k_sparsification}
\end{figure}

\paragraph{Sparsification of the RWR Matrix.}
To tackle the issue of storing in memory large RWR matrices, we explore how the sparsification of the RWR matrix affects the regularization of the model. In particular, we apply a \textit{top-$K$} strategy: for each node, we only keep the $K$ highest RWR weights (in literature there are also efficient methods to directly compute only the top-$K$ RWR weights \cite{Wei_2018}). Figure \ref{fig:gcn_top_k_sparsification} shows how different values of $K$ impact performance on node classification (which usually is the task with the largest graphs). We can see that the addition of the RWRReg term is always beneficial. Furthermore, by taking the \textit{top-$\frac{n}{2}$}, we can reduce the number of entries in the RWR matrix of $\frac{n^2}{2}$ elements, while still obtaining an average \textbf{3.2\%} increment on the accuracy of the model. This strategy then allows the selection of the value of $K$ that best suits the available memory, while still obtaining a high performing model (better than GCN without global structural information injection).

\begin{figure}
\begin{center}
  \subfloat[]{\includegraphics[width=0.42\linewidth]{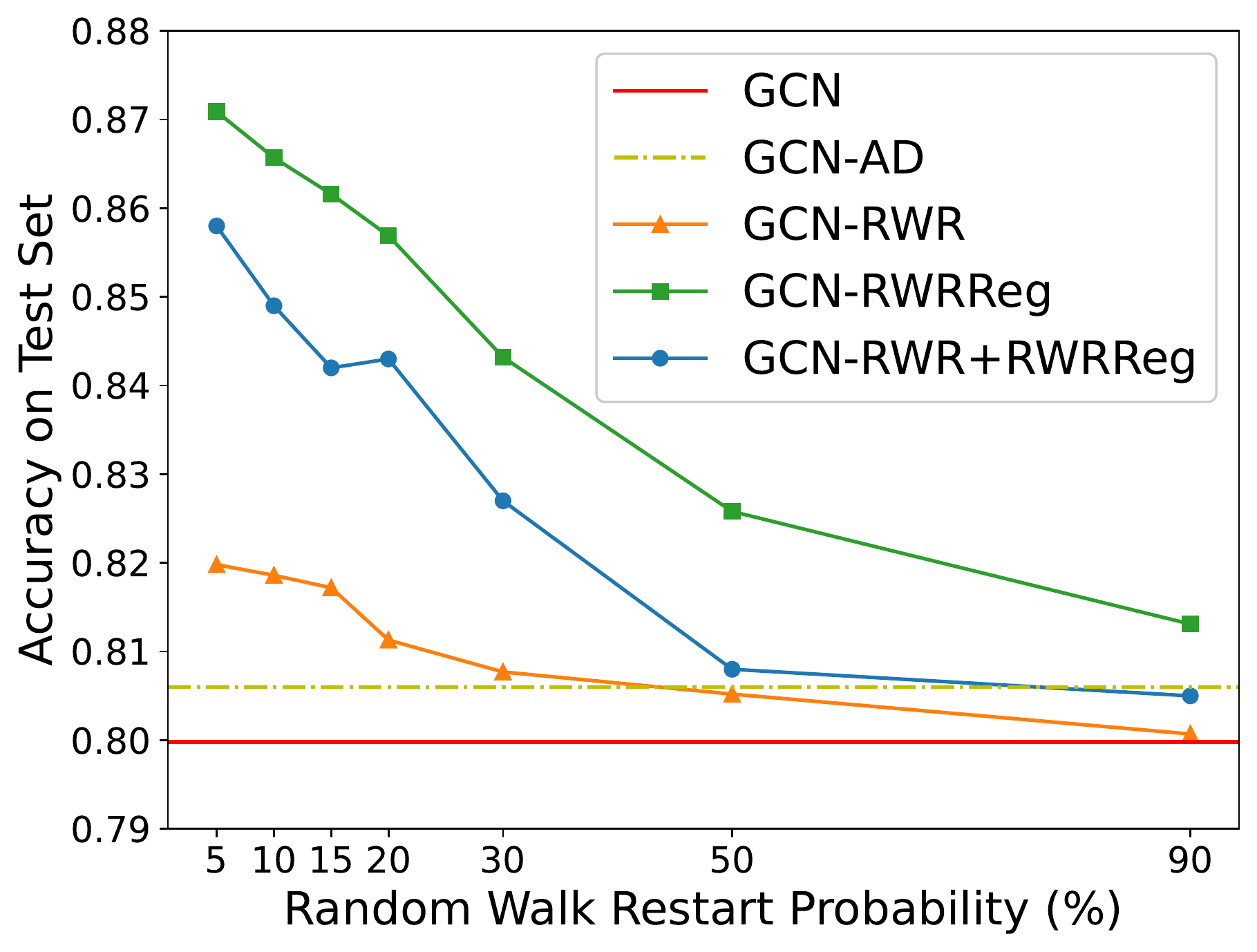}}
  \subfloat[]{\includegraphics[width=0.42\linewidth]{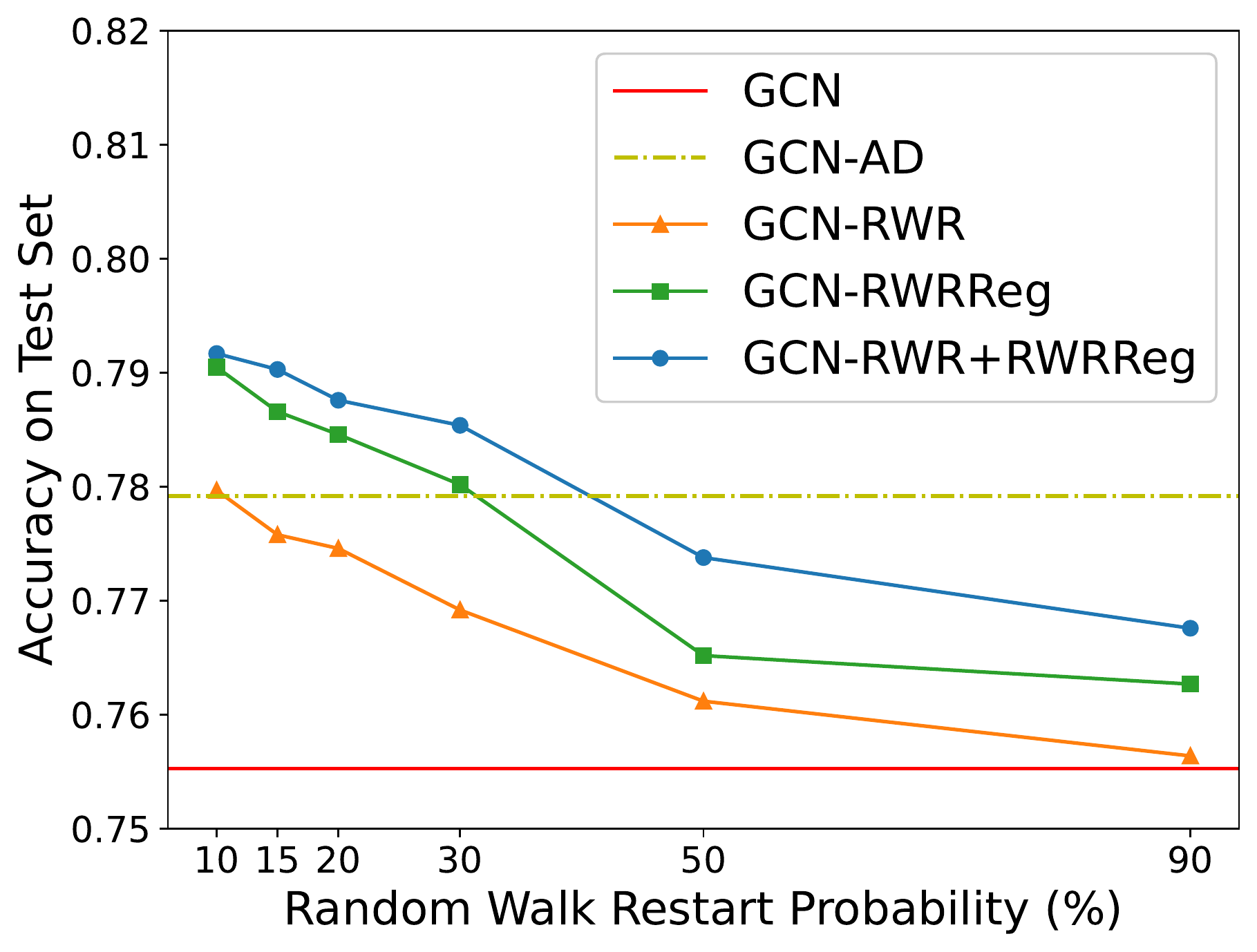}}
\end{center}
 \caption{Accuracy on Cora (a), and on D\&D (b), of GCN without and with the injection of structural information, and for different restart probabilities of RWR.}
  \label{fig:gcn_cora_rws_plt}
\end{figure}

\paragraph{Impact of RWR Restart Probability.}
The use of RWR requires to set the restart probability parameter. We show how performance changes with different restart probabilities. Intuitively, higher restart probabilities might put more much focus on close nodes, as the random walker with frequently return to the starting node. On the other side, lower probabilities allow for more long-range exploration, but may get ``trapped'' into densely connected subgraphs.
Intuitively we would expect lower probabilities to provide more information that is not already available to practical GNNs, and hence lead to higher performance.
Figure \ref{fig:gcn_cora_rws_plt} summarises how the accuracy on node classification (side (a)) and graph classification (side (b)) changes with different restart probabilities\footnote{We did not go below $5\%$ for Cora, and $10\%$ for D\&D for stability reasons in the computation of the RWR coefficients.}. In accordance to our intuition, higher restart probabilities focus on close nodes (and less on distant nodes), and produce lower accuracies. Furthermore, we notice how injecting RWR information is never detrimental to the performance of the model without any injection.

\section{Related Work}\label{S:relwork}
The field of GNNs has become extremely vast, for a thorough review we refer the reader to a recent survey on the subject~\citep{Wu2019ACS}. To the best of our knowledge there are no studies that test if global information regarding the whole graph can significantly impact MPNNs on real-world tasks. However, there are some works that are conceptually related to our approach. 


Several works have taken advantage of RWR in the context of MPNNs. \citet{klicpera_diffusion_2019} use RWR to create a new (weighted) adjacency matrix where message passing is performed. \citet{Li2018DeeperII} use random walks in a co-training scenario to add new nodes for the MPNNs' training set. \citet{Ying_2018} and \citet{Zhang_2019} use random walks to define aggregation neighbourhoods that are not confined to a fixed distance. \citet{AbuElHaija2018NGCNMG} and \citet{mixhop} use powers of the adjacency matrix, which can be considered as random walk statistics, to define neighbourhoods of different scales. \citet{Zhuang_2018} use random walks to define the positive pointwise mutual information (PPMI) matrix and then use it in place of the adjacency matrix in the MPNN formulation. \citet{Klicpera2019PredictTP} use a diffusion strategy based on RWR instead of aggregating information from neighbours. This last work has recently been extended by \citet{Bojchevski2020ScalingGN} to scale to large graphs using RWRs to sample neighbourhoods. We remark how the aforementioned works focus on creating novel MPNN models, while we are interested in studying the impact of global structural information (which MPNNs do not have access to).

\citet{Gao_2019}, and \citet{Jiang2018GraphLR} use regularization techniques to enforce that the embeddings of neighbouring nodes should be close to each other. The first uses Conditional Random Fields, while the second uses a regularization term based on the graph Laplacian. Both approaches only focus on 1-hop neighbours and do not take global information into account.

With regards to the study of the capabilities and weaknesses of GNNs, \citet{Li2018DeeperII} and \citet{Xu2018RepresentationLO} study the over-smoothing problem that appears in Deep-GCN architectures, while \citet{Xu2018HowPA} and \citet{Morris2019} characterize the relation to the Weisfeiler-Leman algorithm. Other works have expressed the similarity with distributed computing~\citep{Sato2019ApproximationRO,Loukas2020What}, and the alignment with particular algorithmic structures~\citep{xu2020what}. These important contributions have advanced our understanding of the capabilities of GNNs, but they do not analyze or quantify the impact of \textit{global} structural information.

Our RWRReg term relies on the computation of the RWR coefficients for every node (for computing the loss function). When dealing with large graphs, there is a vast literature on fast approximations of RWR scores~\citep{Andersen_2006,Tong2006FastRW,Bahmani_2010,Lofgren2015EfficientAF,Wei_2018,Wang_2019}.

Recent work~\citep{micali2016reconstructing} has shown that \emph{anonymous random walks} (i.e., random walks where the global identities of nodes are not known) of fixed length starting at node $u$ are sufficient to reconstruct the local neighborhood within a fixed distance of a node $u$~\citep{micali2016reconstructing}. Subsequently, anonymous random walks have been introduced in the context of learning graph representations~\citep{ivanov2018anonymous}. Such results are complementary to ours, since they assume access to the distribution of \emph{entire walks} of a given length, while our RWR representation only stores information on the probability of ending in a given node. In addition, such works do not provide a connection between RWR and 1-WL.

\section{Conclusions}
\label{S:concl}
Whether \textit{global} structural information (i.e., information that depends on the structure of the whole graph) is needed in GNNs for common tasks on graph-structured data is an open question.
In this work we tackle this question directly at its root. In particular, we identify three strategies to inject \textit{global} structural information into MPNN models, and we quantify 
their impact on popular downstream tasks. Our experiments show that the additional information significantly 
boosts the performance of all considered state-of-the-art models, highlighting and quantifying the imporatance that 
\textit{global} structural information can have on common MPNN applications.
We further discuss a novel practical regularization technique based on RWR, which leads to an average improvement of $5\%$ on all models, 
and is supported by a novel connection between RWR and the 1-Weisfeiler-Leman algorithm. 

\authorcontributions{Conceptualization, D.B. and F.V.; methodology, D.B. and F.V.; software, D.B.; validation, D.B. and F.V.; formal analysis, D.B. and F.V.; investigation, D.B.; resources, F.V.; data curation, D.B.; writing---original draft preparation, D.B. and F.V.; writing---review and editing, D.B. and F.V.; visualization, D.B.; supervision, F.V.; project administration, F.V.; funding acquisition, F.V. All authors have read and agreed to the published version of the manuscript.}

\funding{This work was supported, in part, by MIUR of Italy, 
under PRIN Project n. 20174LF3T8 AHeAD and grant L. 232 (Dipartimenti di Eccellenza), 
and by the University of Padova under project ``SID 2020: RATED-X''.}



\conflictsofinterest{The authors declare no conflict of interest. The funders had no role in the design of the study; in the collection, analyses, or interpretation of data; in the writing of the manuscript, or in the decision to publish the~results.}

\appendixtitles{yes} 
\appendixstart
\appendix

\section{Proof of Proposition 1}\label{appendix_proof}
Given a graph $G=(V,E)$, we define its \emph{$k$-step RWR representation} as the set of vectors $\mathbf{r}_v  = [r_{v,u_1}, \dots, r_{v,u_n}]$, $v\in V$, where each entry $r_{v,u}$ describes the probability that a RWR of length $k$ starting in $v$ ends in $u$.

\begin{Proposition}
Let $G_1=(V_1,E_1)$ and $G_2=(V_2,E_2)$ be two non-isomorphic graphs for which the 1-WL algorithm terminates with the correct answer after $k$ iterations and starting from the labelling of all $1$'s. Then the $k$-step RWR representations of $G_1$ and $G_2$ are different.
\end{Proposition}
\begin{proof}
Consider the WL algorithm with initial labeling given by all 1's. It's easy to see that (i) after $k$ iterations the label of a node $v$ corresponds to the information regarding the degree distribution of the neighborhood of distance $\le k$ from $v$ and (ii) in iteration $i \le k$, the degrees of nodes at distance $i$ from $v$ are included in the label of $v$. 
In fact, after the first iteration, two nodes have the same colour if they have the same degree, as the colour of each node is given by the multiset of the colours of its neighbours (and we start with initial labeling given by all 1's). 
After the second colour refinement iteration two nodes have the same colour if they had the same colour after the first iteration (i.e., have the same degree), and the multisets containing the colours (degrees) of their neighbours are the same. In general, after the $k$-th iteration, two nodes have the same colour if they had the same colour in iteration $k-1$, and the multiset containing the degrees of the neighbours at distance $k$ is the same for the two nodes. Hence, two nodes that have different colours after a certain iteration, will have different colours in all the successive iterations. Furthermore, the colour after the $k$-th iteration depends on the colour at the previous iteration (which ``encodes'' the distribution of degree of neighbours up to distance $k-1$ included), and the multiset of the degrees of neghbours at distance $k$.

Given two non-isomorphic graphs $G_1$ and $G_2$, if the WL algorithm terminates with the correct answer starting from the all $1$'s labelling in $k$ iterations, it means that there is no \emph{matching} between vertices in $V_1$ and vertices in $V_2$ such that matched vertices have the same degree distribution for neighborhoods at distance exactly $k$. Equivalently, any matching $M$ that minimizes the number of matched vertices with different degree distribution has at least one such pair. Now consider one such matching $M$, and let $v \in V_1$ and $w \in V_2$  be vertices matched in $M$ with different degree distributions for neighborhoods at distance exactly $k$. Since $v$ and $w$ have different degree distributions at distance $k$, the number of choices for paths of length $k$ starting from $v$ and $w$ must be different (since the number of choices for the $k$-th edge on the path is different). Therefore, there must be at least a node $u \in V_1$ and a node $z \in V_2$ that are matched by $M$ but for which the number of paths of length $k$ from $v$ to $u$ is different from the number of paths of length $k$ from $w$ to $z$. Since $r_{v,u}$ is proportional to the number of paths of length $k$ from $v$ to $u$, we have that $r_{v,u} \neq r_{w,z}$, that is $\mathbf{r}_v \neq \mathbf{r}_w$. Thus, the  \emph{$k$-step RWR representation} of $G_1$ and $G_2$ are different.
\end{proof}

\section{Model Implementation Details}\label{Appendix_model_details}
We present here a detailed description of the implementations of the models we use in our experimental section. Whenever possible, we started from the official implementation of the authors of each model. Table \ref{tab:model_implementations} contains links to the implementations we used as starting point for the code for our experiments. Our code is available as supplementary material and will be made publicly available after acceptance.

\paragraph{Training Details.} With regards to the training procedure we have that all models are trained with early stopping on the validation set (stopping the training if the validation loss doesn't decrease for a certain amount of epochs), and unless explicitly specified, we use Cross Entropy as loss function for all the classification tasks.

For the task of graph classification we zero-pad the feature vectors of each node to make them all the same length when we inject structural information into the node feature vectors. 

For the task of triangle counting we follow \cite{knyazev2019understanding} and use the one-hot representation of node degrees as node feature vectors to impose some structural information in the network.

\paragraph{Computing Infrastructure.} The experiments were run on a GPU cluster with 7 Nvidia 1080Ti, and on a CPU cluster (when the memory consumption was too big to fit in the GPUs) equipped with 8 cpus 12-Core Intel Xeon Gold 5118 @2.30GHz, with 1.5Tb of RAM.
\newline \newline
In the rest of this Section we go through each model used in our experiments, specifying architecture, hyperparameters, and the position of the node embeddings used for RWRReg.
\begin{table}[]
\caption{\label{tab:model_implementations}Starting model implementations.}
\begin{center}
\begin{tabular}{ l l}
 \toprule 
 \textbf{Model} & \textbf{Implementation} \\
 \midrule
 GCN     \textit{(for node classification)}     & \href{https://github.com/tkipf/pygcn}{github.com/tkipf/pygcn} \\ 
 GCN \textit{(for graph classification)} & \multirow{2}{*}{\href{https://github.com/bknyaz/graph\_nn}{github.com/bknyaz/graph\_nn}}\\
 GCN \textit{(for triangle counting)} & \\
 GraphSage    & \href{https://github.com/williamleif/graphsage-simple/}{github.com/williamleif/graphsage-simple}\\  
 GAT     & \href{https://github.com/Diego999/pyGAT}{github.com/Diego999/pyGAT}\\
 DiffPool & \href{https://github.com/RexYing/diffpool}{github.com/RexYing/diffpool}\\
 $k$-GNN & \href{https://github.com/chrsmrrs/k-gnn}{github.com/chrsmrrs/k-gnn} \\
 \bottomrule
\end{tabular}
\end{center}
\end{table}

\subsection{GCN \textit{(node classification)}}
We use a two layer architecture. The first layer outputs a 16-dimensional embedding vector for each node, and passes it through a ReLu activation, before applying dropout \cite{JMLR:v15:srivastava14a}, with probability $0.5$. The second layer outputs a $c$-dimensional embedding vector for each node, where $c$ is the number of output classes and these vectors are passed through \textit{Softmax} to get the output probabilities for each class. An additional L2-loss is added with a balancing term of $0.0005$. The model is trained using the Adam optimizer \cite{Kingma2015} with a learning rate of 0.01.

We apply the RWRReg on the 16-dimensional node embeddings after the first layer. 

\subsection{GCN \textit{(graph classification)}}
We first have two GCN layers, each one generating a 128-dimensional embedding vector for each node. Then we apply \textit{max}-pooling on the features of the nodes and pass the pooled 128-dimensional vector to a two-layer feed-forward neural network with 256 neurons at the first layer and $c$ at the last one, where $c$ is the number of output classes. A ReLu activation is applied in between the two feed-forward layers, and \textit{Softmax} is applied after the last layer. Dropout \cite{JMLR:v15:srivastava14a} is applied in between the last GCN layer and the feed-forward layer, and in between the feedforward layers (after ReLu), in both cases with probability of 0.1. The model is trained using the Adam optimizer \cite{Kingma2015} with a learning rate of 0.0005.

We apply the RWRReg on the 128-dimensional node embeddings after the last GCN layer. 

\subsection{GCN \textit{(counting triangles)}}
We first have three GCN layers, each one generating a 64-dimensional embedding vector for each node. Then we apply \textit{max}-pooling on the features of the nodes and pass the pooled 64-dimensional vector to a one-layer feed-forward neural network with one neuron. Dropout \cite{JMLR:v15:srivastava14a} is applied in between the last GCN layer and the feed-forward layer with probability of 0.1. The model is trained by minimizing the mean squared error (MSE) and is optimized using the Adam optimizer \cite{Kingma2015} with a learning rate of 0.005.

We apply the RWRReg on the 64-dimensional node embeddings after the last GCN layer. 

\subsection{GraphSage}
We use a two layer architecture. For Cora we sample 5 nodes per-neighbourhood at the first layer and 5 at the second, while on the other datasets we sample 10 nodes per-neighbourhood at the first layer and 25 at the second. Both layers are composed of \textit{mean-aggregators} (i.e., we take the mean of the feature vectors of the nodes in the sampled neighbourhood) that output a 128-dimensional embedding vector per node. After the second layer these embeddings are multiplied by a learnable matrix with size $128 \times c$, where $c$ is the number of output classes, giving thus a $c$-dimensional vector per-node. These vectors are passed through \textit{Softmax} to get the output probabilities for each class. The model is optimized using Stochastic Gradient Descent with a learning rate of 0.7.

We apply the RWRReg on the 128-dimensional node embeddings after the second aggregation layer.

\subsection{GAT}
We use a two layer architecture. The first layer uses an 8-headed attention mechanism that outputs an $8$-dimensional embedding vector per-node. LeakyReLu is set with slope $\alpha = 0.2$. Dropout \cite{JMLR:v15:srivastava14a} (with probability of 0.6) is applied after both layers. The second layer outputs a $c$-dimensional vector for each node, where $c$ is the number of classes, and before passing each vector through \textit{Softmax} to obtain the output predictions, the vectors are passed through an Elu activation \cite{Clevert2016}. An additional L2-loss is added with a balancing term of $0.0005$. The model is optimized using Adam \cite{Kingma2015} with a learning rate of 0.005.

We apply the RWRReg on the 8-dimensional node embeddings after the first attention layer. A particular note needs to be made for the training of GATs: we found that naively implementing the RWRReg term on the node embeddings in between two layers brings to an exploding loss as the RWRReg term grows exponentially at each epoch. We believe this happens because the attention mechanism in GATs allows the network to infer that certain close nodes, even 1-hop neighbours, might not be important to a specific node and so they shouldn't be embedded close to each other. This clearly goes in contrast with the RWRReg loss, since 1-hop neighbours always have a high score. We solved this issue by using the attention weights to scale the RWR coefficients at each epoch (we make sure that gradients are not calculated for this operation as we only use them for scaling). This way the RWRReg penalizations are in accordance with the attention mechanism, and are still encoding long-range dependencies.

\subsection{DiffPool}
We use a 1-pooling architecture. The initial node feature matrix is passed through two (one to obtain the assignment matrix and one for node embeddings) 3-layer GCN, where each layer outputs a 20-dimensional vector per-node. Pooling is then applied, where the number of clusters is set as 10\% of the number of nodes in the graph, and then another 3-layer GCN is applied to the pooled node features. Batch normalization \cite{Ioffe2015} is added in between every GCN layer. The final graph embedding is passed through a 2-layer MLP with a final \textit{Softmax} activation. An additional L2-loss is added with a balancing term of $10^{-7}$, together with two pooling-specific losses. The first enforces the intuition that nodes that are close to each other should be pooled together and is defined as: $\mathcal{L}_{LP} = \Vert \boldsymbol{A}^{(l)}, \boldsymbol{S}^{(l)^{\intercal}}  \boldsymbol{S}^{(l)} \Vert_{F}$, where $\Vert \cdot \Vert_{F}$ is the Frobenius norm, and $ \boldsymbol{S}^{(l)}$ is the assignment matrix at layer $l$. The second one encourages the cluster assignment to be close to a one-hot vector, and is defined as: $\mathcal{L}_{E} = \frac{1}{n} \sum_{i=1}^{n} H(\boldsymbol{S}_{i,:})$, where $H$ is the entropy function. However, in the implementation available online, the authors do not make use of these additional losses. We follow the latter implementation. The model is optimized using Adam \cite{Kingma2015} with a learning rate of 0.001.

We apply the RWRReg on the 20-dimensional node embeddings after the first 3-layer GCN (before pooling). We tried applying it also after pooling on the coarsened graph, but the fact that this graph could change during training yields to poor results.

\subsection{$k$-GNN}
We use the hierarchical 1-2-3-GNN architecture (which is the one showing the highest empirical results). First a 1-GNN is applied to obtain node embeddings, then these embeddings are used as initial values for the 2 GNN (1-2-GNN). The embeddings of the 2-GNN are then used as initial values for the 3-GNN (1-2-3-GNN). The 1-GNN applies 3 graph convolutions, while 2-GNN and the 3-GNN apply 2 graph convolutions. Each convolution outputs a 64-dimensional vector and is followed by an Elu activation \cite{Clevert2016}. For each $k$, node features are then globally averaged and the final vectors are concatenated and passed through a three layer MLP. The first layer outputs a 64-dimensional vector, while the second outputs a 32-dimensional vector, and the third outputs a $c$-dimensional vector, where $c$ is the number of output classes. To obtain the final output probabilities for each class, \textit{log(Softmax)} is applied, and the negative log likelihood is used as loss function. After the first and the second MLP layers an Elu activation \cite{Clevert2016} is applied, furthermore, after the first MLP layer dropout \cite{JMLR:v15:srivastava14a} is applied with probability 0.5. The model is optimized using Adam \cite{Kingma2015} with a learning rate of 0.01, and a decaying learning rate schedule based on validation results (with minimum value of $10^{-5}$). 

We apply the RWRReg on the 64-dimensional node embeddings after the $1$-GNN. We were not able to apply it also after the 2-GNN and the 3-GNN, as it would cause out-of-memory issues with our computing resources.

\section{Datasets}\label{appendix_Dataset_statistics}
We briefly present here some additional details about the datasets used for our experimental section. Table \ref{tab:node_classification_dataset_stats} summarizes the datasets for node classification, while Table \ref{tab:graph_classification_dataset_stats} presents information about the datasets for graph classification and triangle counting. The node classification datasets are available at \url{https://linqs.soe.ucsc.edu/data}, while the graph classification and the triangle counting at \url{https://chrsmrrs.github.io/datasets/}.
\begin{table}[]
\caption{\label{tab:node_classification_dataset_stats}Node classification dataset statistics.}
\begin{center}
\begin{tabular}{ l c c c c c}
 \toprule 
 \textbf{Dataset} & \textbf{Nodes} & \textbf{Edges} & \textbf{Classes} & \textbf{Features} & \textbf{Label Rate} \\
 \midrule 
 Cora          & 2708    & 5429   & 7 & 1433 & 0.052\\ 
 Pubmed    & 19717 & 44338 & 3 & 500 & 0.003\\  
 Citeseer    & 3327   & 4732   & 6 & 3703 & 0.036\\
 \bottomrule 
\end{tabular}
\end{center}
\end{table}

\begin{table}[]
\caption{\label{tab:graph_classification_dataset_stats}Graph classification and triangle counting dataset statistics.}
\begin{center}
\begin{tabular}{ l c c c c c}
 \toprule 
 \textbf{Dataset} & \textbf{Graphs} & \textbf{Classes} & \textbf{Avg. \# Nodes} & \textbf{Avg. \# Edges}\\
 \midrule 
 ENZYMES          & 600    & 6   & 32.63 & 62.14\\ 
 D\&D    & 1178 & 2 & 284.32 & 715.66\\  
 PROTEINS    & 1113   & 2   & 39.1 & 72.82\\
 \midrule
 TRIANGLES & 45000 & 10 & 20.85 & 32.74\\
 \bottomrule
\end{tabular}
\end{center}
\end{table}

\section{Fast Implementation of the Random Walk with Restart Regularization}\label{appendix:compute_rwr}
Let $\boldsymbol{H}$ be the matrix containing the node embeddings, and $\boldsymbol{S}$ be the matrix with the RWR statistics. We are interested in the following quantity
\[ \mathcal{L}_{\text{\textit{RWRReg}}} = \sum_{i,j} \boldsymbol{S}_{i,j} \lvert \lvert \boldsymbol{H}_{i, :} - \boldsymbol{H}_{j, :} \rvert \rvert^{2} \]
To calculate it in a fast way (specially when using GPUs) we use the following procedure. Let us first define the following matrices:
\begin{align*}
\boldsymbol{\hat S} &= n \times n \text{ symmetric matrix with } \boldsymbol{\hat S}_{i,j} = 
\begin{cases}
\boldsymbol{S}_{i,j}+\boldsymbol{S}_{j,i} &  \text{for } i \neq j\\
\boldsymbol{S}_{i,j} & \text{for } i = j\\
\end{cases}\\
\boldsymbol{D} &= n \times n \text{ diagonal matrix with } D_{i,i} = \sum_{j} \boldsymbol{\hat S}_{i, j}\\
\boldsymbol{\Delta} &= \boldsymbol{D} - \boldsymbol{\hat S}
\end{align*}
We then have
\[
\mathcal{L}_{\text{\textit{RWRReg}}} = \sum_{i,j} \boldsymbol{S}_{i,j} \lvert \lvert \boldsymbol{H}_{i, :} - \boldsymbol{H}_{j, :} \rvert \rvert^{2} 
= \sum_{i} \boldsymbol{H}_{:, i}^\intercal \boldsymbol{\Delta} \boldsymbol{H}_{:, i} 
= Tr \left( \boldsymbol{H}^\intercal \boldsymbol{\Delta} \boldsymbol{H} \right)
\]
Where $Tr(\cdot)$ is the trace of the matrix. Note that $\boldsymbol{H}_{:, i}^\intercal$ is the $i$-th column of $\boldsymbol{H}$, transposed, so its size is $1 \times n$.

\section{Empirical Analysis of the Random Walk with Restart Matrix}\label{S:rw_analysis}
We now analyse the RWR matrix to justify the use of RWR for the encoding of global structural information. We consider the three node classification datasets (see Section \ref{S:eval} of the paper), as this is the task with the largest input graphs, and hence where this kind of information seems more relevant. 

\begin{figure}[h!]
\centering
  \subfloat[]{\includegraphics[height=1.3in]{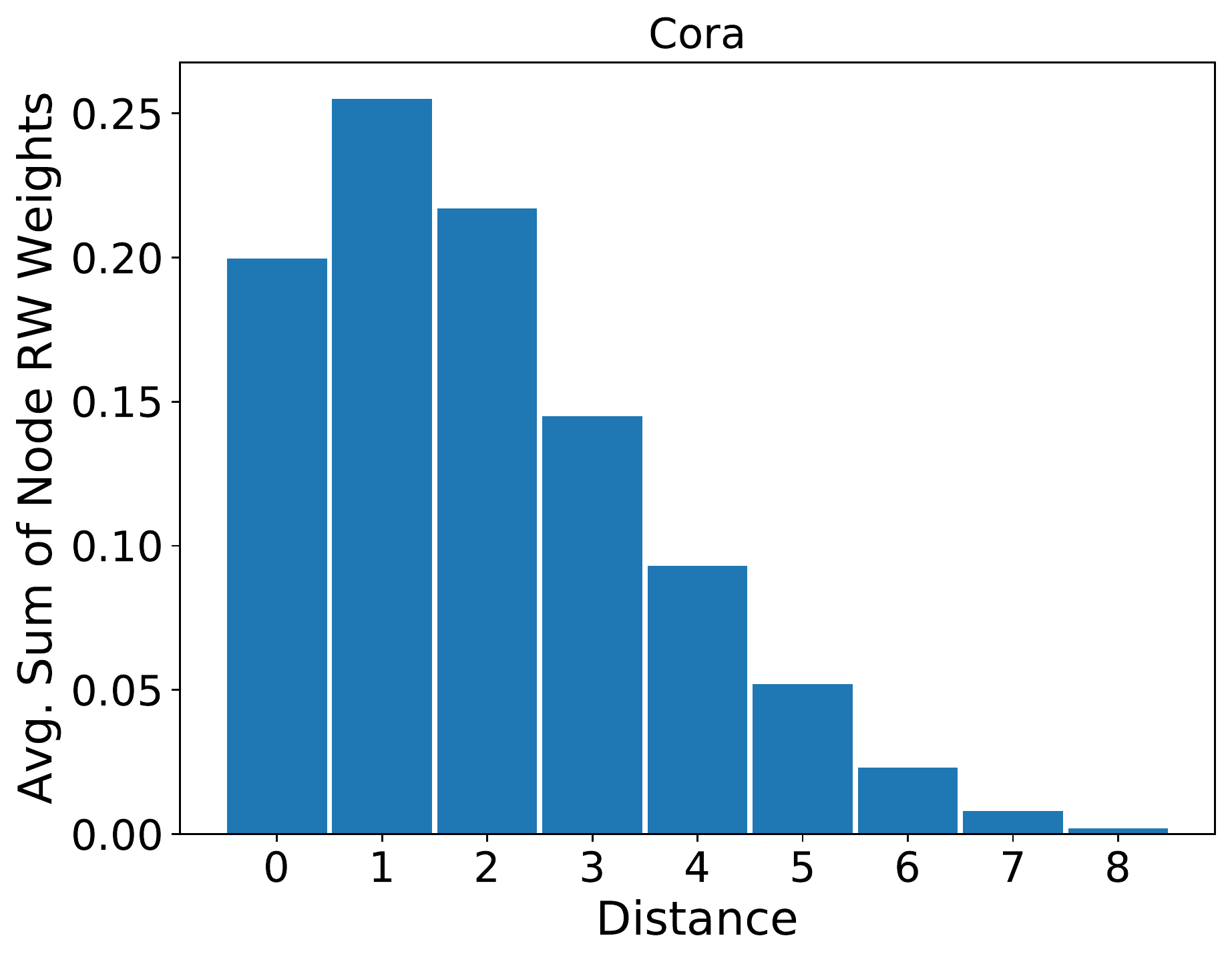}}
  \subfloat[]{\includegraphics[height=1.3in]{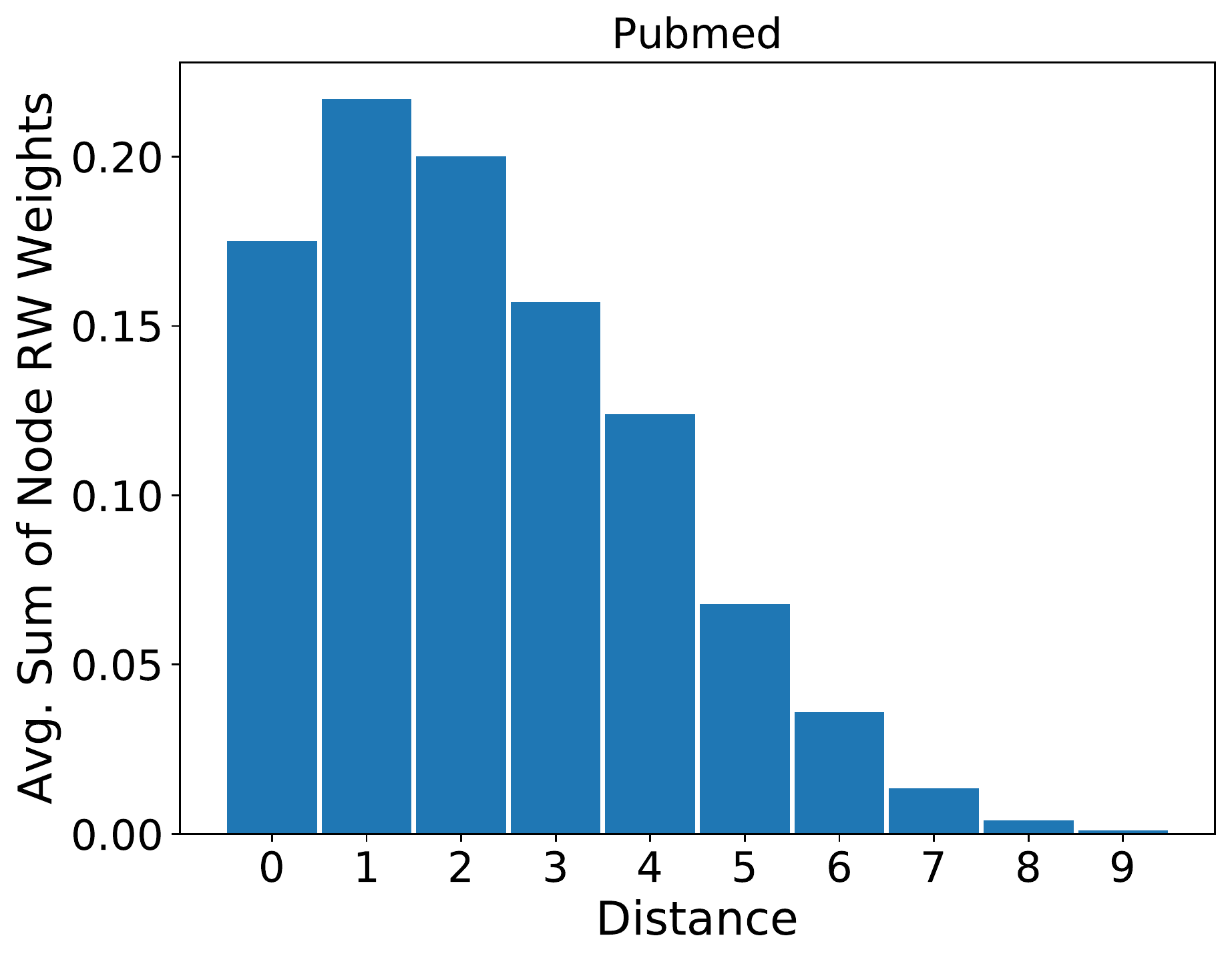}}
  \subfloat[]{\includegraphics[height=1.3in]{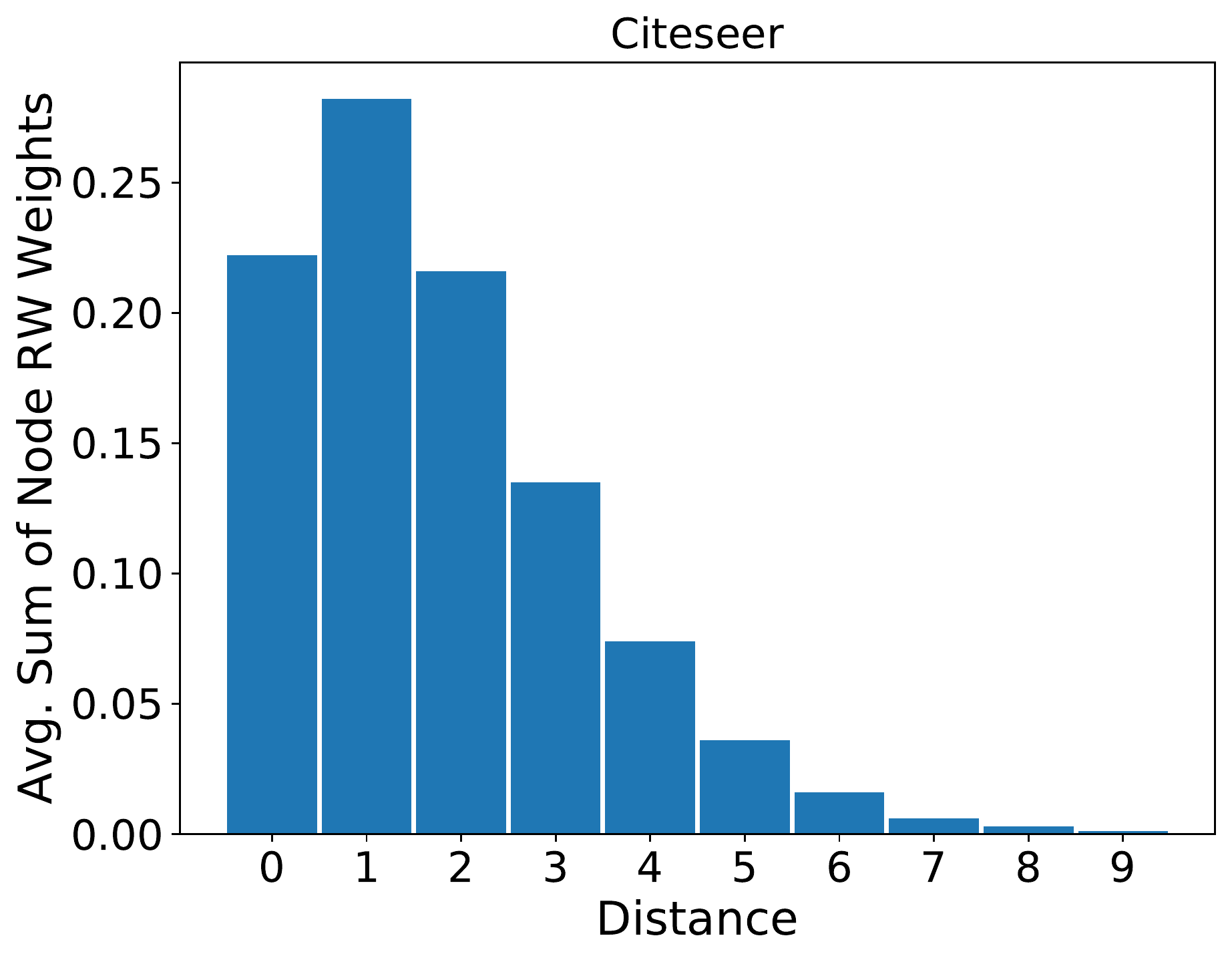}}
  \caption{Average distribution of the RWR weights at different distances for the node classification datasets. Distance zero indicates the weight that a node assigns to itself.}
  \label{fig:rw_weights_at_distances}
\end{figure}

\begin{wraptable}{r}{6.0cm}
\caption{\label{tab:avg_datset_kt_values}Average and standard deviation, over all nodes, of Kendall Tau-b values measuring the non-trivial relationships between nodes captured by the RWR weights.}
\vspace{0.2cm}
\begin{tabular}{ l c c c }
\toprule
  \textbf{Dataset} & \textbf{Average Kendall Tau-b}\\
 \midrule
 Cora            & $0.729 \pm 0.082$\\ 
 Pubmed      & $0.631 \pm 0.057$\\  
 Citeseer      & $0.722 \pm 0.171$\\
 \bottomrule
\end{tabular}
\end{wraptable}
We first consider the distribution of the RWR\footnote{We consider RWR, with a restart probability of $0.15$, as done for the experimental evaluation of our proposed technique.} weights at different distances from a given node. In particular, for each node, we take the sum of the weights assigned to the 1-hop neighbours, the 2-hop neighbours, and so on. We then take the average, over all nodes, of the sum of the RWR weights at each hop. We discard nodes that belong to connected components with diameter $\le 4$, and we only plot the values for the distances that have an average sum of weights higher than $0.001$. Plots are shown in Figure \ref{fig:rw_weights_at_distances}. We notice that the RWR matrix contains information that goes beyond the immediate neighbourhood of a node. In fact, we see that approximately $90\%$ of the weights are contained within the 6-hop neighbourhood, with a significant portion that is not contained in the 2-hop neighbourhood usually accessed by MPNN models.

Next we analyse if RWR capture some non-trivial relationships between nodes. In particular, we investigate if there are nodes that are far from the starting node, but receive a higher weight than some closer nodes. To quantify this property we use the Kendall Tau-b\footnote{We use the Tau-b version because the elements in the sequences we analyze are not all distinct.} measure (\cite{10.2307/2332303}). 
In more detail, for each node $v$ we consider the sequence $rw^{(v)}$ where the $i$-th element is the weight that the RWR from node $v$ has assigned to node $i$: $rw^{(v)}[i] = S_{v,i}$. We then define the sequence $drw^{(v)}$ such that $drw^{(v)} [j] = \text{\textit{dist}}(v, f_{sort\_weights}(j, rw^{(v)})) $, where \textit{dist(x, y)} is the shortest path distance between node $x$ and node $y$, and $f_{sort\_weights}(j, rw^{(v)})$ is the node with the $j$-th highest RWR weight in $rw^{(v)}$.
Intuitively, if the RWR matrix isn't capable of capturing non-trivial relationship we would have that $drw^{(v)}$ is a sorted list (with repetitions). By comparing $drw^{(v)}$ with its sorted version with the Kendall Tau-b rank, we obtain a value between 1 and $-1$ where 1 means that the two sequences are identical, and $-1$ means that one is the reverse of the other. Table \ref{tab:avg_datset_kt_values} presents the results, averaged over all nodes, on the node classification datasets. These results show that while there is a strong relation between the information provided by RWR and the distance between nodes, there is information in the RWR that is not captured by shortest path distances.

\begin{figure}[h!]
\centering
  \includegraphics[height=4in]{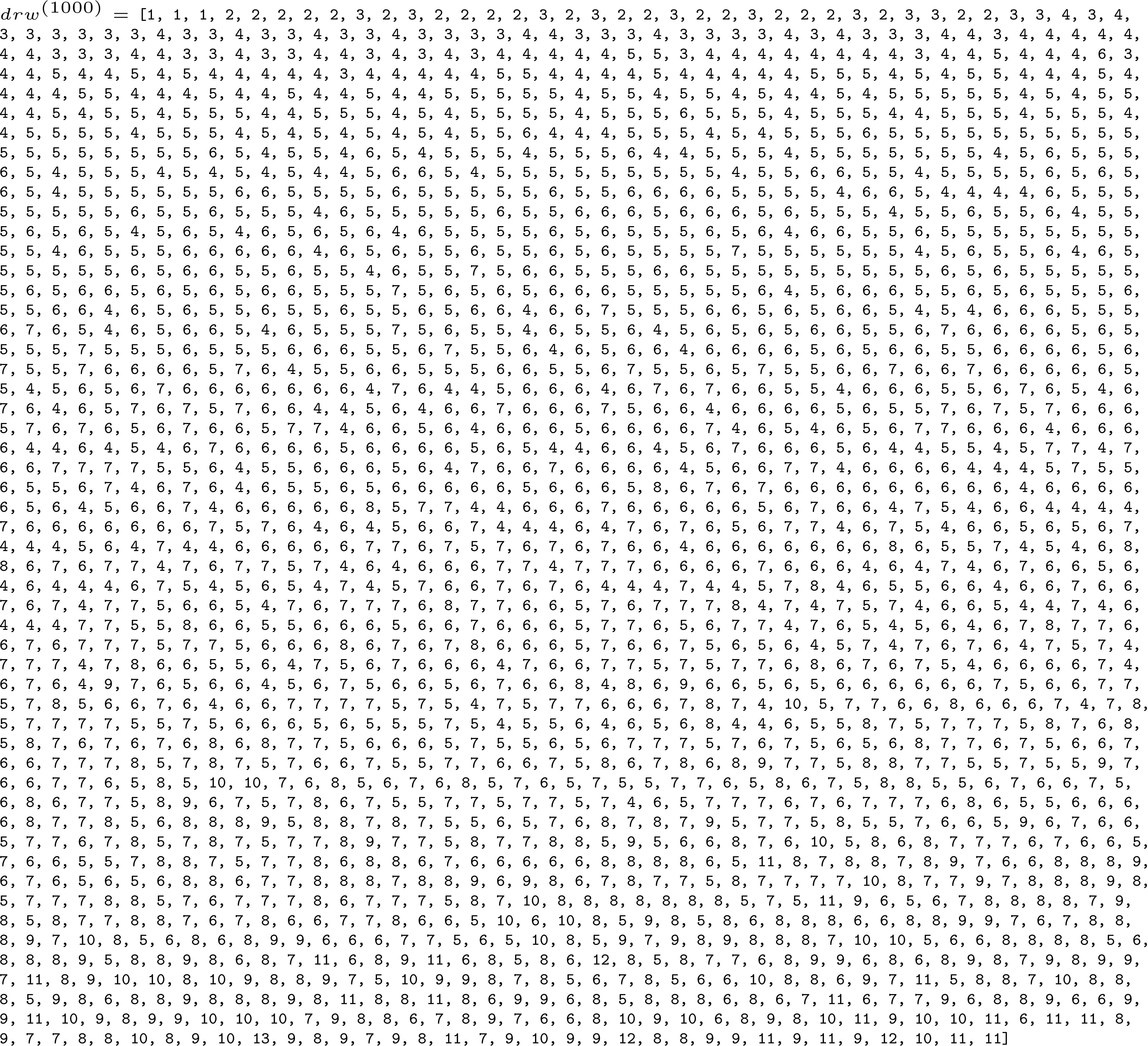}
  \caption{$drw^{(v)}$ sequence for the $1000$-th node in Cora.}
  \label{fig:drw_seq}
\end{figure}

As an example of the non-trivial relationships encoded by RWR, Figure \ref{fig:drw_seq} presents a $drw^{(v)}$ sequence taken from a node in Cora. This sequence obtains a Kendall Tau-b value of $0.591$. We can observe that for distances greater than 1, we already have some non-trivial relationships. In fact, we observe some nodes at distance 3 that receive a larger weight than nodes at distance 2. There are many other interesting non-trivial relationships, for example we notice that some nodes at distance 7, and some at distance 11, obtain a higher weight than some nodes at distance 5.

\end{paracol}
\reftitle{References}

\externalbibliography{yes}
\bibliography{citations}

%

\end{document}